\documentclass[journal,twoside,web]{ieeecolor}
\usepackage{tmi}
\usepackage{cite}
\usepackage{nicematrix}      
\usepackage{amsmath,amssymb,amsfonts}
\usepackage{graphicx}
\usepackage{textcomp}
\usepackage[lined,ruled,vlined]{algorithm2e}
\definecolor{mycommentcolor}{RGB}{0, 128, 0}

\SetCommentSty{mycomment}
\SetKwComment{Comment}{/* }{ */}
\usepackage[colorlinks=true, citecolor=blue, linkcolor=blue, urlcolor=blue]{hyperref}
\usepackage{longtable}
\newlength\savewidth\newcommand\shline{\noalign{\global\savewidth\arrayrulewidth
  \global\arrayrulewidth 1.25pt}\hline\noalign{\global\arrayrulewidth\savewidth}}
\usepackage{booktabs}
\usepackage{bbding}
\usepackage{latexsym}
\usepackage{framed,multirow}
\usepackage{threeparttable}
\usepackage{xspace}
\usepackage{cite}
\newcommand{\std}[1]{{}}
\usepackage{algpseudocode}
\usepackage{bm} 
\usepackage{bbm}

\newcommand{\ie}{\textit{i}.\textit{e}.}
\newcommand{\eg}{\textit{e}.\textit{g}.}

\newcommand{\eat}[1]{}

\newcommand{\tabref}[1]{Table~\ref{#1}}
\newcommand{\figref}[1]{Fig.~\ref{#1}}

\newcommand{\secref}[1]{Sec.~\ref{#1}}

\newcommand{\vct}[1]{\boldsymbol{#1}} 

\newcommand{\methodname}{{Brain-WM}\xspace}
\newcommand{\revision}[1]{{#1}}

\def\BibTeX{{\rm B\kern-.05em{\sc i\kern-.025em b}\kern-.08em
    T\kern-.1667em\lower.7ex\hbox{E}\kern-.125emX}}
\begin{document}
\title{\methodname: Brain Glioblastoma World Model}

\author{
Chenhui Wang, 
Boyun Zheng, 
Liuxin Bao, 
Zhihao Peng,
Peter Y. M. Woo,\\
Hongming Shan,~\IEEEmembership{Senior Member, IEEE},
Yixuan Yuan,~\IEEEmembership{Senior Member, IEEE}
\thanks{This work was supported by NSFC/RGC Joint Research Scheme N\_CUHK4126/25, Innovation and Technology Fund Mainland-Hong Kong Joint Funding Scheme MHP/173/24 (Corresponding authors: Yixuan Yuan; Hongming Shan)}
\thanks{
C. Wang and H. Shan are with the Institute of Science and Technology for Brain-inspired Intelligence, Fudan University, Shanghai 200433, China (e-mail: chenhuiwang21@m.fudan.edu.cn; hmshan@ieee.org)}
\thanks{
B. Zheng, Z. Peng, and Y. Yuan are with the Department of Electronic Engineering, The Chinese University of Hong Kong, Hong Kong SAR, China (e-mail: byzheng@link.cuhk.edu.hk; zhihao.peng@cityu.edu.hk; yxyuan@ee.cuhk.edu.hk)}
\thanks{L. Bao is with the School of Automation, Hangzhou Dianzi University, Hangzhou 310018, China (e-mail: lxbao@hdu.edu.cn).}
\thanks{P. Woo is with the Department of Neurosurgery, Prince of Wales Hospital, Hong Kong, SAR, China (e-mail: peterwoo@surgery.cuhk.edu.hk)}
}

\maketitle
\begin{abstract}
Precise prognostic modeling of glioblastoma (GBM) under varying treatment interventions is essential for optimizing clinical outcomes. 
While generative AI has shown promise in simulating GBM evolution, existing methods typically treat interventions as static conditional inputs rather than dynamic decision variables.
Consequently, they fail to capture the complex, reciprocal interplay between tumor evolution and treatment response.
To bridge this gap, we present \methodname, a pioneering brain GBM world model that unifies next-step treatment prediction and future MRI generation, thereby capturing the co-evolutionary dynamics between tumor and treatment.
Specifically, \methodname encodes spatiotemporal dynamics into a shared latent space for joint autoregressive treatment prediction and flow-based future MRI generation. 
Then, instead of a conventional monolithic framework, \methodname adopts a novel Y-shaped Mixture-of-Transformers (MoT) architecture. This design structurally disentangles heterogeneous objectives, successfully leveraging cross-task synergies while preventing feature collapse. 
Finally, a synergistic multi-timepoint mask alignment objective explicitly anchors latent representations to anatomically grounded tumor structures and progression-aware semantics.
Extensive validation on internal and external multi-institutional cohorts demonstrates the superiority of \methodname, achieving 91.5\% accuracy in treatment planning and SSIMs of 0.8524, 0.8581, and 0.8404 for FLAIR, T1CE, and T2W sequences, respectively.
Ultimately, \methodname offers a robust clinical sandbox for optimizing patient healthcare. The source code is
made available at \url{https://github.com/thibault-wch/Brain-GBM-world-model}.

\end{abstract}

\begin{IEEEkeywords}
Glioblastoma (GBM), World Model, Treatment, Magnetic Resonance Imaging (MRI)
\end{IEEEkeywords}

\section{Introduction}
\label{sec:intro}
Glioblastoma (GBM) is the most   aggressive primary malignant brain tumor in adults, characterized by diffuse infiltration and a near-universal recurrence rate that results in a median survival of only 11--15 months~\cite{stupp2005radiotherapy,woo2023patterns}.
Although practical guidelines recommend maximal safe resection followed by chemoradiotherapy and adjuvant temozolomide, real-world adherence varies significantly~\cite{shergalis2018current}.
Factors such as advanced age, frailty, and rapid progression often force deviations into heterogeneous second-line regimens or premature discontinuation. 
Consequently, patients with similar baseline characteristics or tumor molecular profiles  often experience markedly divergent clinical trajectories, underscoring the critical need to model tumor evolution across varying treatment interventions.

\begin{figure}[!tp]
	\centering
\includegraphics[width=.9\linewidth]{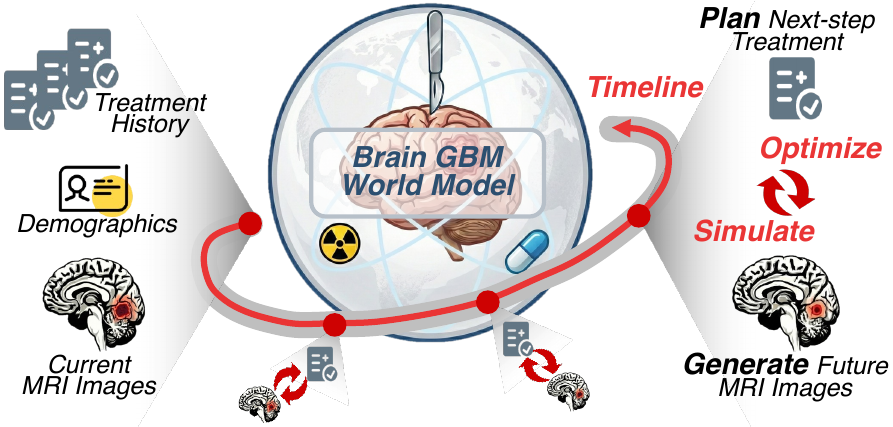}
\caption{\textbf{Conceptual overview of \methodname.} \methodname establishes the first brain GBM world model by encoding spatiotemporal dynamics from treatment history, demographics, and multi-modal MRI to unify next-step clinical planning and future tumor morphology synthesis, thereby capturing the co-evolutionary dynamics between treatment and tumor.}
\label{fig:concept}
\end{figure}

Current approaches to modeling GBM evolution broadly fall into two categories: analytical biophysical frameworks  and data-driven generative models.
Biophysical frameworks utilize mathematical equations to simulate tumor growth and tissue displacement based on known biological principles~\cite{clatz2005realistic,hogea2008image}. While highly interpretable, these frameworks often struggle with extreme patient heterogeneity  and require intensive manual parameter tuning~\cite{lipkova2019personalized}.
Conversely, generative models have demonstrated significant potential to synthesize high-fidelity tumor growth patterns directly from longitudinal medical imaging~\cite{petersen2019deep,leclercq2025pre,liu2025treatment}.
Despite these gains, such models remain predominantly unidirectional and passive, treating interventions as static conditional inputs rather than dynamic variables. Consequently, they fail to capture the reciprocal feedback loop between  tumor evolution and treatment response.

To overcome these limitations, we introduce the first world model~\cite{ha2018world,lecun2022path} tailored for neuro-oncology, framing longitudinal GBM progression as a continuous, interactive sequence of clinical actions and pathological states. 
While world models excel in robotics and autonomous systems~\cite{jiang2025irl,li2025recogdrive,cen2025rynnvla}, their translation to the human brain presents unique challenges. First, a significant semantic gap exists between discrete, low-dimensional treatment planning and continuous, high-dimensional MRI generation.
Naively entangling these heterogeneous modalities inevitably triggers feature interference, disrupting the reciprocal feedback loop necessary for optimal clinical forecasting.
Second, anatomical hallucinations frequently emerge during future MRI generation due to an absence of strict neuroanatomical constraints. To ensure biological plausibility, the model must transcend pixel-level optimization to explicitly capture the morphological characteristics of distinct GBM subregions, \eg~the peritumoral edema, enhancing tumor, and necrotic core.

In this work, we introduce \methodname, a brain GBM world model designed to unify next-step treatment planning with future MRI generation, as illustrated in \figref{fig:concept}.
Moving beyond previous static paradigms,  \methodname explicitly captures the co-evolutionary dynamics between tumor and treatment: predicted pathological trajectories inform treatment planning, while treatment decisions constrain the biological plausibility of the simulated images. 
To address the aforementioned semantic gap, we bypass a conventional monolithic design in favor of an advanced Y-shaped Mixture-of-Transformers (MoT).
This architectural innovation elegantly disentangles two fundamentally heterogeneous objectives: autoregressive modeling for discrete treatment forecasting and continuous flow-based generation for MRI synthesis.
Furthermore, to ensure anatomical consistency, we propose a synergistic multi-timepoint mask alignment (MM-Align) objective. 
By anchoring shared intermediate representations to time-specific masks, the model prioritizes structural tumor priors before determining treatment plans or synthesizing fine-grained appearances. Ultimately, \methodname provides a robust in-silico sandbox for clinicians to explore counterfactual scenarios and optimize personalized patient healthcare.

The contributions of this work are summarized as follows.
\begin{enumerate}
\item We propose \methodname, a novel brain GBM world model that unifies autoregressive treatment prediction and flow-based future MRI generation by encoding spatiotemporal dynamics into a shared latent space.
\item We develop a novel Y-shaped MoT architecture that structurally disentangles distinct planning and imaging objectives, leveraging cross-task synergies while preventing feature collapse in task-specific subspaces.
\item We design a synergistic multi-timepoint mask alignment objective that anchors shared representations to current and future tumor masks, regularizing the latent space with anatomically grounded, progression-aware semantics for superior cross-task performance.
\item Extensive validation on  internal and external cohorts shows the superior performance and generalizability of \methodname over state-of-the-art (SOTA) methods in both  treatment planning and future MRI generation.
\end{enumerate}

\section{Related Works}

\subsection{Deep Learning for GBM}
\label{related:gbm}
Most deep learning studies on GBM focus on discriminative endpoints like segmentation and survival stratification~\cite{menze2014multimodal,chen2025autoregressive}. 
To improve precision, recent frameworks inject richer clinical priors into visual backbones: 
multi-granularity automated and editable prompt learning (MGAEPL)~\cite{sun2025mgaepl} leverages tumor grade for boundary refinement, while multimodal graph~\cite{sun2024cross} and attention-based models~\cite{zhong2025thread} incorporate demographic and transcriptomic profiles for more robust prognosis.
However, they largely produce static outputs, rather than modeling longitudinal tumor dynamics.
To address this gap, emerging generative methods~\cite{leclercq2025pre,liu2025treatment} have begun to study treatment-conditioned tumor evolution, but they typically cover 1-2 treatment categories and are often validated on relatively small cohorts.
In contrast, \methodname models treatment as a dynamic decision variable through the unification of treatment intent planning and tumor evolution simulation, and is systematically validated across multi-centric cohorts.

\subsection{World Model}
\label{related:world}
Current world models primarily serve two functions: learning implicit representations to capture environmental mechanisms, \eg~joint embedding predictive architecture~\cite{lecun2022path}, and predicting future states for robotics and autonomous driving~\cite{jiang2025irl,li2025recogdrive,cen2025rynnvla}. While preliminary adaptations like the medical world model (MedWM)~\cite{yang2025medical} introduce this paradigm to healthcare, it inherently isolates treatment planning from future image synthesis. Crucially, adapting world models to GBM is hindered by the challenge of modeling intricate tumor evolution within restrictive brain anatomy.
To bridge this gap, we propose \methodname, the first dedicated brain world model. \methodname unifies treatment planning and future MRI synthesis within a cohesive architecture; pivotal to this integration is a novel synergistic multi-timepoint mask alignment to enforce anatomical consistency and progression awareness, thereby advancing precision neuro-oncology.

\section{Methodology}
This section first introduces the problem formulation in \secref{sec:overview} and the proposed multi-modal tokenizers in \secref{sec:tok}. Next, we present the Y-shaped unified MoT architecture and the synergistic multi-timepoint mask alignment in \secref{sec:arc} and \secref{sec:mask}, respectively. Finally, we detail the objective function used to optimize \methodname in \secref{sec:object}.

\begin{figure*}[!htbp]
	\centering
\includegraphics[width=.9\linewidth]{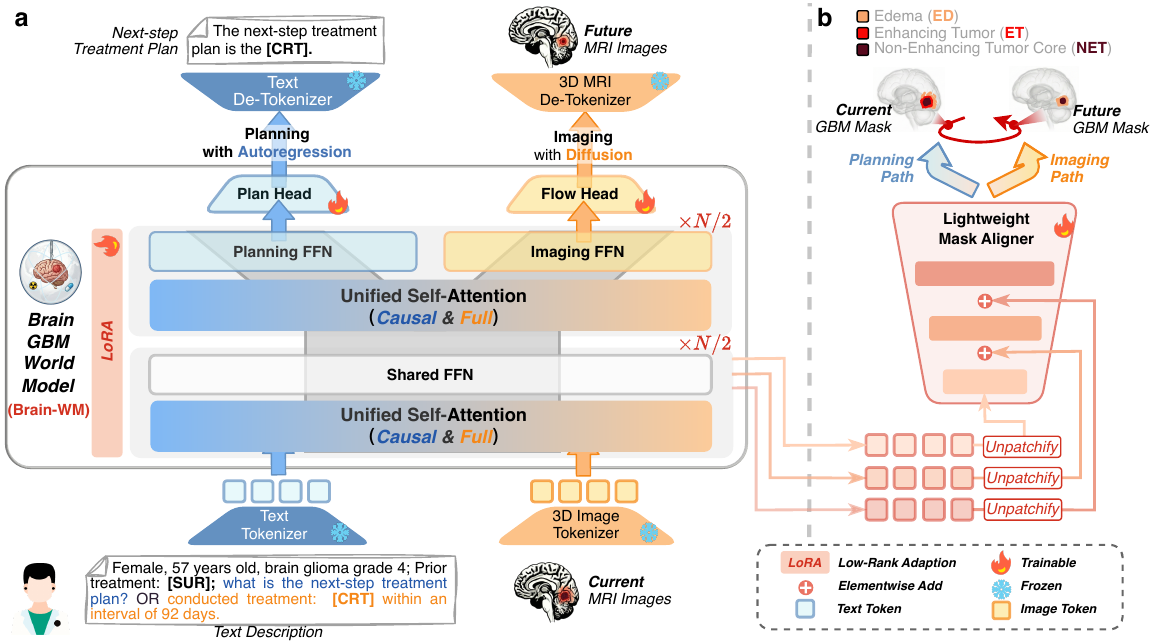}
 \caption{\textbf{Overview of our \methodname.} (\textbf{a}) \methodname features a unified Y-shaped Mixture-of-Transformers (MoT) backbone that bifurcates to optimize distinct objectives: autoregressive treatment planning and flow-based future MRI generation. (\textbf{b}) A multi-timepoint mask alignment objective ensures the learned representations remain anatomically grounded and task-consistent across time.
 }
\label{fig:detail}
\end{figure*}

\begin{figure}[!t]
	\centering
\includegraphics[width=.95\linewidth]{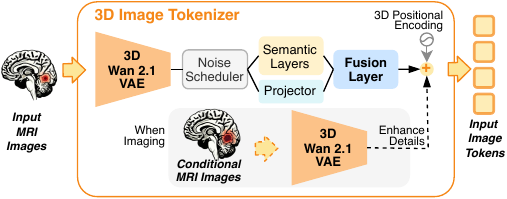}
 \caption{\textbf{Illustration of the volumetric MRI tokenizer.}}
\label{fig:imgtok}
\end{figure}

\subsection{Problem Formulation}
\label{sec:overview}
The primary objective of \methodname is to learn a unified GBM world model, denoted as $\mathcal{M}_\psi$, that encapsulates the latent dynamics of tumor progression to support two complementary tasks: (i) a \textit{planning branch} that recommends the next-step treatment plan, and (ii) an \textit{imaging branch} that forecasts future MRI states under a specified treatment intervention, as shown in \figref{fig:detail}(a).

We formulate the patient state at a clinical time point $p$ as a multi-modal tuple $\mathcal{S}_p = (\vct{x}_{p}, \vct{c}_{p})$. Here, $\vct{x}_p$ represents the multi-parametric MRI images (FLAIR, T1CE, and T2W), while $\vct{c}_{p}$ denotes the \textit{textual clinical context}, encompassing the cumulative history of past treatment plans $\vct{c}^\text{plan}_{1:p}$ and static demographic descriptors $\vct{c}^\text{demo}$.
Conditioned on the current patient state $\mathcal{S}_p$, the model $\mathcal{M}_\psi$ is trained to jointly optimize two generative objectives: (a) autoregressively predicting the next-step treatment plan $\hat{\vct{c}}_{p+1}^\text{plan}$, and (b) synthesizing the future MRI images $\hat{\vct{x}}_{p+1}$ given a specific candidate next-step treatment plan and a forecasting interval $\tau$.
Formally, we instantiate the unified GBM world model $\mathcal{M}_\psi$ as:
\begin{equation}
\mathcal{M}_\psi:
\begin{cases}
\hat{\vct{c}}_{p+1}^\text{plan} 
= \mathcal{M}_\psi^{\text{plan}}\!\big(\mathcal{S}_p\big), \\[0.25em]
\hat{\vct{x}}_{p+1}
= \mathcal{M}_\psi^{\text{img}}\!\big(\mathcal{S}_p, \vct{c}_{p+1}^\text{plan}, \tau\big),
\end{cases}
\end{equation}
where $\mathcal{M}_\psi^{\text{plan}}$ and $\mathcal{M}_\psi^{\text{img}}$ denote the specialized planning and imaging heads of the world model, respectively. 
Crucially, the imaging branch is explicitly conditioned on the specified treatment plan $\vct{c}_{p+1}^\text{plan}$ (either predicted by the model or manually defined), enforcing a causal dependency between treatment action and tumor evolution.

\subsection{Multi-modal Tokenizers}
\label{sec:tok}
To unify treatment planning and imaging, we encode multi-parametric 3D MRI images and clinical narratives as an interleaved sequence of tokens, structuring the input as:
\[
\texttt{[BOS]}\,\texttt{[BOI]}\,\{\texttt{MRI}\}\,\texttt{[EOI]}\,\{\texttt{TXT}\}\,\texttt{[EOS]},
\]
where \texttt{[BOS]} and \texttt{[EOS]} mark sequence boundaries, while \texttt{[BOI]} and \texttt{[EOI]} enclose multi-parametric 3D MRI images. The block $\{\texttt{TXT}\}$ encodes demographics and treatment history, followed by either a query for treatment planning or a specified treatment for future MRI synthesis.

\subsubsection{Clinical Text Tokenizer}
To explicitly capture the full spectrum of the patient treatment trajectory, we augment the standard vocabulary with five special planning tokens: 1) \texttt{[SUR]} for surgical tumor resection; 2) \texttt{[CRT]} for concurrent chemoradiotherapy; 3) \texttt{[RT]} for standalone radiotherapy; 4) \texttt{[TMZ]} for adjuvant temozolomide chemotherapy; and 5) \texttt{[AM]} for active monitoring or participation in clinical trials.

The $\{\texttt{TXT}\}$ block is instantiated using a template structure. For example, history is encoded as: 
\emph{``Female, 57 years old, brain glioma grade 4. Prior treatment:} \texttt{[SUR]}\emph{.''}
Subsequently, the prompt diverges by task: the planning branch utilizes a fixed query
\emph{``What is the next-step treatment plan?''},
whereas the imaging branch conditions on a specified treatment plan
\emph{``Conducted treatment:} \texttt{[SUR]} + \texttt{[CRT]}\emph{ over an interval of 92 days.''}
To preserve the original linguistic manifold, we freeze the pre-trained vocabulary weights, fine-tuning only the embeddings of the above five special planning tokens.

\subsubsection{Volumetric MRI Tokenizer} 
We adapt the unified image tokenizer from Show-o2~\cite{xie2025show} for volumetric processing of the block $\{\texttt{MRI}\}$, as shown in~\figref{fig:imgtok}.
At each time point $p$, multi-sequence MRI inputs $\vct{x}_p$ (FLAIR, T1CE, T2W) are concatenated channel-wise. 
The resulting volume is encoded via the 3D  Wan2.1 VAE~\cite{wan2025wan}, integrated with distilled SignLIP~\cite{zhai2023sigmoid} semantic layers and low-level projectors.
This architecture yields a dual-purpose representation where the planning branch utilizes deterministic visual tokens, while the imaging branch operates on noise-augmented latents. 
To preserve the intrinsic anatomical topology, we superimpose learnable 3D absolute positional embeddings, anchoring each token within the volumetric coordinate system. 
Notably, the imaging branch is conditioned on the current MRI latent $\vct{x}_{p}$ to enhance fine-grained structural details in future forecasting.

\subsection{Y-shaped Mixture-of-Transformers (MoT)}
\label{sec:arc}
While most existing vision-language foundation models~\cite{li2023llava,zhang2024vision} typically rely on a monolithic transformer optimized for a single objective, our framework necessitates balancing the distinct regimes of  treatment planning and volumetric imaging.
To leverage cross-task synergies while preserving task-specific representation subspaces, we propose a Y-shaped MoT architecture.

Given a unified token sequence $\vct{H}_0 \in \mathbb{R}^{L \times d}$ comprising concatenated imaging and planning tokens of length $L$ and hidden dimension $d$, we employ a transformer with $N$ layers. 
The initial $N_{\text{sha}} = N/2$ layers serve as a shared feature encoder to capture multimodal correlations, while the subsequent $N_{\text{spe}} = N/2$ layers bifurcate into task-specific branches.
Throughout the network, we utilize a \emph{unified} self-attention mechanism across all layers:
\begin{equation}
    \hat{\vct{H}}_\ell
    = \mathrm{UniAttn}\bigl(\vct{H}_{\ell-1}; \mathbf{M}\bigr)
      + \vct{H}_{\ell-1}, \quad \ell = 1, \dots, N,
\end{equation}
where $\mathrm{UniAttn}(\cdot)$ denotes multi-head self-attention~\cite{vaswani2017attention} parameterized jointly for both modalities. 

To govern token visibility, the mask $\mathbf{M} \in \mathbb{R}^{L \times L}$ enforces causal dependencies for next-step treatment planning while enabling full context for future MRI generation.
Let $\mathcal{I}_{\text{text}}$ denote the indices of text tokens in $\vct{H}_0$, we define:
\begin{equation}
    \mathbf{M}_{ij} =
    \begin{cases}
        -\infty, & i,j \in \mathcal{I}_{\text{text}},\ j > i, \\[2pt]
        0,       & \text{otherwise}.
    \end{cases}
\end{equation}
This configuration ensures autoregressive causality within the textual domain while permitting bidirectional full attention within the visual domain.

The Y-shaped topology is realized via task-conditional Feed-Forward Networks (FFNs). For a specific task $\mathrm{tsk} \in \{\mathrm{plan}, \mathrm{img}\}$, the layer-wise FFN update is defined as:
\begin{equation}
    \vct{H}_\ell^{(\text{tsk})} =
    \begin{cases}
        \hat{\vct{H}}_\ell^{(\text{tsk})} +
        \mathrm{FFN}_\ell^{\text{shared}}\bigl(\hat{\vct{H}}_\ell^{(\text{tsk})}\bigr),
        & \ell \le N_{\text{sha}}, \\[4pt]
        \hat{\vct{H}}_\ell^{(\text{tsk})} +
        \mathrm{FFN}_\ell^{(\text{tsk})}\bigl(\hat{\vct{H}}_\ell^{(\text{tsk})}\bigr),
        & \ell > N_{\text{sha}},
    \end{cases}
    \label{eq:y_mot_ffn}
\end{equation}
where $\mathrm{FFN}_\ell^{\text{shared}}$ is shared across tasks in the initial layers, while $\mathrm{FFN}_\ell^{(\text{tsk})}$ becomes task-specific in the subsequent layers.

Finally, the model employs a plan head for autoregressive treatment prediction and a flow head for future MRI generation. The plan head is implemented as a linear layer, while the flow head consists of a transformer stack with adaptive layer normalization modulated by diffusion time steps.

\subsection{Synergistic Multi-timepoint Mask Alignment}
\label{sec:mask}
While the Y-shaped MoT architecture bifurcates for distinct objectives, the shared backbone lacks the supervisory guidance to capture fine-grained tumor morphology and leverage inter-task synergy.
To address this, we introduce a synergistic multi-timepoint mask alignment objective, termed MM-Align, inspired by representation alignment (REPA)~\cite{yu2024representation}, as shown in \figref{fig:detail}(b).
This auxiliary supervision anchors the shared latent space to clinically meaningful structures, ensuring that features remain anatomically grounded and semantically coherent across temporal domains.
Specifically, we dynamically supervise the shared representations using the \emph{current} tumor mask for the planning task and the \emph{future} tumor mask for the imaging task, enforcing a unified structural regularization that promotes robust representation learning.

\subsubsection{Mask Aligner Architecture}
We attach a lightweight \emph{mask aligner} to the shared transformer to promote structure-aware and time-relevant representation learning.
Let $\mathbf{H}^{(i)}$ denote intermediate image tokens from the shared transformer at depth  $i$, we aggregate multi-scale features $\{\mathbf{H}^{(i)}\}_{i\in\mathcal{I}}$ for a predefined index set  $\mathcal{I}$ via top--down pyramid fusion~\cite{lin2017feature} to predict voxel-wise logits $\mathbf{Z}$, which are converted into a probabilistic segmentation map $\hat{\mathbf{m}}$ via a softmax function:
\begin{equation}
\hat{\mathbf{m}}=\mathrm{Softmax}(\mathbf{Z}),
\end{equation}
where $\mathbf{Z}\in\mathbb{R}^{B\times C\times D\times H\times W}$, and $C$ represents four segmentation classes,  comprising the background and three GBM sub-regions: edema (ED), enhancing tumor (ET), and the non-enhancing tumor core (NET).

\subsubsection{Dynamic Multi-timepoint Supervision}
To accommodate the dual objectives of our framework, we employ dynamic supervision targets. Let $\mathbf{m}^{\mathrm{cur}}$ and $\mathbf{m}^{\mathrm{fut}}\in\{0,1\}^{B\times C\times D\times H\times W}$ denote the one-hot ground-truth masks for the current and future timepoints, respectively. 
The active supervision mask $\mathbf{m}$ is selected based on the task context: $\mathbf{m}=\mathbf{m}^{\mathrm{cur}}$ for the planning and $\mathbf{m}=\mathbf{m}^{\mathrm{fut}}$ for the imaging branch.

We optimize a compound loss that consists of a voxel-wise focal loss~\cite{lin2017focal} and a region-based Dice loss~\cite{milletari2016v}.
For a voxel $v\in\Omega$ with ground-truth class $c_v$, the focal loss is:
\begin{equation}
\mathcal{L}_{\mathrm{focal}} = - \frac{1}{|\Omega|} \sum_{v \in \Omega} \big(1 - \hat{m}_{v,c_v}\big)^{\gamma} \log\big(\hat{m}_{v,c_v}\big),
\end{equation}
where $\gamma$ is the focusing parameter. The total focal loss $\mathcal{L}_\mathrm{focal}$ is the mean over all voxels.

Concurrently, for each tumor sub-region $r\in\mathcal{R}_C$ (ED/ET/NET), we minimize the Dice loss:
\begin{equation}
\mathcal{L}_{\mathrm{dice}} = \frac{1}{|\mathcal{R}_C|} \sum_{r \in \mathcal{R}_C} \left( 1 - \frac{2\langle \hat{\mathbf{m}}_{r}, \mathbf{m}_{r}\rangle}{\|\hat{\mathbf{m}}_{r}\|_{1} + \|\mathbf{m}_{r}\|_{1}} \right),
\end{equation}
where $\hat{\mathbf{m}}_{r}$ and $\mathbf{m}_{r}$ are the flattened prediction and label maps for region $r$, respectively. Then, the total Dice loss $\mathcal{L}_\mathrm{dice}$ is the mean over all regions.

Finally, we formulate the unified alignment objective as:
\begin{equation}
\mathcal{L}_{\mathrm{align}}
=\alpha(t)\Big(\mathcal{L}_{\mathrm{focal}}+\mathcal{L}_{\mathrm{dice}}\Big),
\end{equation}
where $\alpha(t)$ acts as a time-dependent weighting coefficient. For the planning task, we fix $\alpha(t)=1$ to yield a deterministic trajectory. For the imaging task, $\alpha(t)$ is modulated by the flow-matching noise schedule, assigning lower weights to high-noise steps and higher weights to low-noise steps to ensure reliable structural alignment.
At inference, MM-Align can be detached, but retaining it provides more interpretable  outputs that offer  meaningful visual cues to support clinical review.

\subsection{Objective Function}
\label{sec:object}
Our \methodname provides two capabilities:  autoregressive treatment planning and treatment-conditioned tumor forecasting. 
Both tasks are supervised via a  synergistic multi-timepoint alignment objective $\mathcal{L}_{\mathrm{align}}$, which regularizes the representation to remain anatomically consistent and progression-aware.

During training, we employ an alternating optimization strategy, randomly assigning each mini-batch to either the planning or imaging branch. 
We rely on the patient state $\mathcal{S}_p = (\vct{x}_{p}, \vct{c}_{p})$ at clinical time $p$, where the textual context $\vct{c}_p$ aggregates  treatment history and demographics.

For treatment planning, we formulate the planning task as autoregressive next-token prediction. Conditioned on the current state $\mathcal{S}_p$, the model minimizes the negative log-likelihood of the next-step treatment plan $\vct{c}^\text{plan}_{p+1}$, regularized by the current-timepoint mask alignment term:
\begin{equation}
\mathcal{L}_{\mathrm{plan}}
= -\log p\Big(\vct{c}_{p+1}^{\mathrm{plan}} \,\big|\, 
\mathcal{S}_p\Big)
+ \mathcal{L}_{\mathrm{align}}.
\end{equation}

For future MRI generation, we adopt a latent flow-matching framework to synthesize the future MRI $\vct{x}_{p+1}$.
Let $\vct{z}_{1}=\mathcal{E}(\vct{x}_{p+1})$ be the target latent encoded by the volumetric encoder $\mathcal{E}(\cdot)$ and  $\vct{z}_0\!\sim\!\mathcal{N}(\mathbf{0},\mathbf{I})$ the source noise.
We define the forward process as a linear interpolation from noise to data:
\begin{equation}
\vct{z}_{t}=t\vct{z}_{1}+(1-t)\vct{z}_{0},\quad t\sim\mathcal{U}(0,1).
\end{equation}
In this formulation, the proposed model $\mathcal{M}_\psi$ learns to predict the constant velocity field that transports the probability density from $\vct{z}_0$ to $\vct{z}_1$.
The prediction is conditioned on the current noisy latent $\vct{z}_{t}$, the patient state $\mathcal{S}_p$, the specified treatment plan $\vct{c}^\text{plan}_{p+1}$, and the timestep $t$:
\begin{equation}
\hat{\vct{v}}_{t}
= \mathcal{M}^{\mathrm{img}}_{\psi}\!\left(\vct{z}_{t},\,\mathcal{S}_p,\,\vct{c}^\text{plan}_{p+1},\,t\right).
\end{equation}
We optimize the flow-matching objective by regressing the targeted direction $\vct{v}_t= \frac{d\vct{z}_t}{dt} =\vct{z}_1-\vct{z}_0$, augmented by the future-timepoint mask-alignment term:
\begin{equation}
\mathcal{L}_{\mathrm{img}}
= \lambda_{\mathrm{img}}\,
\!\left\|\hat{\vct{v}}_{t}-\vct{v}_t\right\|_2^2
+ \mathcal{L}_{\mathrm{align}},
\end{equation}
where $\lambda_{\mathrm{img}}$ balances the generation objective.

At inference, the planning head yields the next-step treatment token $\hat{\vct{c}}_{p+1}^{\mathrm{plan}}$ in a single forward pass. 
For future MRI generation, we generate the future MRI by sampling the initial latent $\vct{z}_0$ from a Gaussian prior and iteratively evolving it with an Euler solver~\cite{chen2018neural} with 50 timesteps.
At each timestep, the model predicts the velocity field conditioned on the current patient state $\mathcal{S}_p$ and the specified plan $\vct{c}_{p+1}^\mathrm{plan}$ to update the latent trajectory. Finally, the terminal latent $\vct{z}_1$ is decoded to obtain the future MRI.

\section{Experiments}
\subsection{Experiment Setup}
\subsubsection{Datasets}
To ensure robust model generalization across diverse clinical settings, we curated our data from multiple institutions. 
The internal cohort was aggregated from three public datasets: LUMIERE~\cite{suter2022lumiere}, MU-Glioma Post~\cite{mahmoud2025mu}, and UCSF-ALPTDG~\cite{fields2024university}, yielding 527 subjects with 1,659 longitudinal timepoints.
To maximize temporal diversity, we employed a dense pairing strategy, treating every distinct timepoint pair within a subject as an independent instance. This yielded 3,252 pairs; we used an 8:2 subject-level split for training and internal validation, such that all derived pairs from the same subject were confined to a single split.
For external evaluation, we applied the same pairing protocol to an independent validation cohort of 61 subjects (128 timepoints) from the RHUH-GBM~\cite{cepeda2023rio} and UCSD-PTGBM~\cite{gagnon2026university} datasets, deriving 72 data pairs.
This external cohort offered a complementary distribution of clinical states, providing a rigorous test for domain generalization: while the internal cohort spans the full spectrum of the GBM treatment pathway, RHUH-GBM is enriched for surgery-related timepoints, whereas UCSD-PTGBM primarily comprises cases under active monitoring. Each timepoint included 3D MRI sequences (FLAIR, T1CE, T2W) and tumor masks.

\subsubsection{Preprocessing} Treatment plans and temporal intervals were organized from the official clinical metadata using ChatGPT-5.1~\cite{achiam2023gpt}, followed by manual verification to guarantee data fidelity.
All MRI scans were processed using a unified pipeline prior to model training. The protocol comprised four standardized steps~\cite{wang2024joint,wang2024hope}: (1) skull-stripping to remove non-brain tissue, (2) spatial registration to a common SRI24 template space~\cite{rohlfing2010sri24} to align anatomical structures, (3) cropping a fixed 3D volume of interest ($192\times192\times140$ voxels) centered on the brain to remove the non-informative background, and (4) per-subject intensity preprocessing that removes outliers and rescales intensities to $[-1,1]$ for stable optimization.

\subsubsection{Implementation Details} Our \methodname builds on the pretrained weights of Show-o2~\cite{xie2025show}, a unified multimodal model that uses Qwen2.5-1.5B-Instruct~\cite{qwen2.5} as the language backbone.
Our network comprises 28 transformer layers, structured as 14 shared layers between the planning and generation branches and 14 task-specific layers. 
Specifically, intermediate features from the shared layers at index set $\mathcal{L}=\{4,10,14\}$ are extracted and fed into the mask aligner.
We optimized the model using AdamW~\cite{loshchilov2018decoupled} in bfloat16 mixed precision via DeepSpeed, employing decoupled learning rates ($4\times10^{-5}$ for LoRA adapters and $2\times10^{-5}$ for the flow head, projection layers, and planning tokens) with a warmup schedule of 50 steps and a weight decay of 0.01.
Following established configurations in unified multimodal models~\cite{bagel,xie2025show}, we set the weight $\lambda_{\mathrm{img}} = 4$.
Standard 3D augmentations (flips, translations, and zooming)~\cite{cardoso2022monai} were applied during training.

\begin{table*}[!tp]
\centering
\caption{\textbf{Quantitative evaluation of treatment planning performance on internal and external cohorts.} Evaluated baselines are categorized into four types: (\textsc{I}) Open-source general, (\textsc{II}) Closed-source general, (\textsc{III}) Medical domain-specific, and (\textsc{IV}) Task-specific models. The best and second-best results are highlighted in  \textbf{bold} and \underline{underline}, respectively.}
\begin{NiceTabular*}{1.\textwidth}{@{\extracolsep{\fill}} lc|cccc|cccc @{}}

\shline
  \multirow{2}{*}{{Methods}}&\multirow{2}{*}{{Type}}&  \multicolumn{4}{c}{\textit{Internal Cohort}}&
\multicolumn{4}{c}{\textit{External Cohort}}\\
&&Accuracy $\uparrow$&F$1$-Score $\uparrow$&Specificity $\uparrow$&Precision $\uparrow$&Accuracy $\uparrow$&F$1$-Score $\uparrow$&Specificity $\uparrow$&Precision $\uparrow$\\
\midrule
Qwen3-VL 4B~\cite{bai2025qwen3vltechnicalreport} &\multicolumn{1}{c}{\multirow{3}{*}{\textsc{I}}} 
&20.1  &27.7  & 80.2 &52.9  &48.2  & 52.6 &87.0  & \textbf{94.4} \\
InterVL3.5 4B~\cite{wang2025internvl3} && 30.2 & 28.4 & 80.0 & 31.3 & 61.1 & 55.7 & 90.8 & 68.0 \\
DeepSeek V3.2~\cite{liu2025deepseek} && 32.4 & 40.5 & 84.6 & 76.8 & 27.8 & 24.1 & 84.0 & 46.9  \\
\hline

Claude 4.5&\multicolumn{1}{c}{\multirow{3}{*}{\textsc{II}}} & 29.3 & 40.2 & 85.5 &\textbf{91.1} & 27.8 & 24.1 & 84.0 & 46.9  \\
Gemini 2.5 Pro~\cite{comanici2025gemini} && 32.2 & 38.8 & 84.3 & 67.8 & 37.0 & 38.3 & 85.2 & 66.1 \\
GPT 5.1~\cite{achiam2023gpt} &&36.1 & 45.1 & 85.8 & 82.1 & 44.4 & 48.9 & 88.3 &  70.2 \\
\hline

MedVLM-R1 2B~\cite{pan2025medvlm} &\multicolumn{1}{c}{\multirow{3}{*}{\textsc{III}}} & 20.5 & 25.9 & 80.7 & 46.9 & 37.0 & 34.9 & 86.3 & 37.7 \\
Hulu-Med 4B~\cite{jiang2025hulu} && 24.6 & 27.5 & 82.2 & 50.9 & 35.2 & 33.4 & 85.7 & 56.6 \\
Linshu 7B~\cite{xu2025lingshu} &&32.7  & 43.0 & 85.5 & 79.4 & 48.2 & 52.1 & 89.6 & 75.9  \\

\hline
AliFuse~\cite{chen2024alifuse} &\multicolumn{1}{c}{\multirow{3}{*}{\textsc{IV}}} &81.6  &79.8  &89.8  &86.3  &73.3  &75.2 &93.1  &76.9  \\
Hyperfusion~\cite{duenias2025hyperfusion} &&85.6  &80.0  &91.8  &82.1  & 79.3 &79.4  &93.8  &80.4  \\
\rowcolor{cyan!10}\textbf{\methodname (Ours)} & &\textbf{91.5}  &\textbf{90.4}  &\textbf{95.8}  &\underline{90.7} &\textbf{83.5}  &\textbf{85.0} &\textbf{94.1}  &\underline{87.2}   \\

\shline
\end{NiceTabular*} 
\label{tab:compare_all_models}
\end{table*}

\begin{table*}[!tp]
\centering
\caption{\textbf{Quantitative evaluation of future MRI generation on internal and external cohorts.} Reconstruction fidelity is measured via NMSE (\%), PSNR (dB), and SSIM ($\times10^{-2}$). The best results are highlighted in \textbf{bold}.}
\label{tab:comp_gene}
\begin{NiceTabular*}{1\textwidth}
{@{\extracolsep{\fill}}l|ccc|ccc|ccc}
\shline
\multirow{2}{*}{{Methods}}
&
  \multicolumn{3}{c}{{FLAIR}} &
  \multicolumn{3}{c}{{T1CE}} &
  \multicolumn{3}{c}{{T2W}} \\
  & NMSE $\downarrow$ & PSNR  $\uparrow$ & SSIM  $\uparrow$
  & NMSE $\downarrow$ & PSNR $\uparrow$ & SSIM  $\uparrow$
  & NMSE  $\downarrow$ & PSNR  $\uparrow$ & SSIM  $\uparrow$ \\
\midrule
\multicolumn{10}{c}{\textit{Internal Cohort}} \\
\midrule
LE-DUNet~\cite{lee2024synthesizing}           &11.54\std{0.21}  &21.35\std{2.40}  &79.81\std{0.20}  &11.63\std{0.21} &21.32\std{2.37}  &80.16\std{0.24}  & 13.34\std{0.18} &20.97\std{1.92}  &77.39\std{0.18}  \\
LE-GAN~\cite{zhang2025language}       &\,\,\,8.86\std{0.12}  &22.50\std{2.30}  &80.85\std{0.09}  &\,\,\,8.11\std{0.11}  &22.90\std{2.49}  &81.26\std{0.11}  &\,\,\,9.55\std{0.09}  &22.41\std{1.73}  &78.00\std{0.06}  \\
IL-CLDM~\cite{ou2024image}        &\,\,\,6.93\std{0.09}  &23.61\std{2.62}  &83.52\std{0.10}  &\,\,\,6.39\std{0.10}  &24.05\std{3.38}  &83.78\std{0.10}  &\,\,\,7.31\std{0.09}  &23.66\std{2.43}  &81.62\std{0.11} \\
LDM-GBM~\cite{leclercq2025pre}    &\,\,\,6.64\std{0.09}&24.00\std{3.35}  &83.94\std{0.11}  &\,\,\,6.14\std{0.09}    &24.26\std{3.71}  &84.52\std{0.15}  &\,\,\,6.77\std{0.09} &24.07\std{3.13}  &82.53\std{0.12}  \\
MedWM~\cite{yang2025medical} &\,\,\,6.59\std{0.10}  &24.23\std{3.85}  & 84.15\std{0.19} &\,\,\,6.11\std{0.09}  &24.34\std{3.52}  &84.65\std{0.22} &\,\,\,6.68\std{0.07}  &24.19\std{3.49}  &82.83\std{0.17}  \\    
TaDiff~\cite{liu2025treatment} &\,\,\,6.42\std{0.10}  &24.37\std{3.77}  &84.43\std{0.15}  &\,\,\,5.96\std{0.11}  &24.45\std{4.29}  &84.95\std{0.19}  &\,\,\,6.35\std{0.09}  &24.40\std{3.54}  &83.30\std{0.14}  \\   
\cellcolor{cyan!10}\textbf{\methodname (Ours)}
& \cellcolor{cyan!10} \,\,\,\textbf{5.93\std{0.11}}& \cellcolor{cyan!10}\textbf{25.05\std{3.41}} & \cellcolor{cyan!10} \textbf{85.24\std{0.14}}
& \cellcolor{cyan!10}
\,\,\,\textbf{5.46\std{0.09}} & \cellcolor{cyan!10}\textbf{25.29\std{3.54}} & \cellcolor{cyan!10}\textbf{85.81\std{0.18}}
& \cellcolor{cyan!10}\,\,\,\textbf{5.98\std{0.08}} & \cellcolor{cyan!10} \textbf{25.17\std{3.09}} & \cellcolor{cyan!10} \textbf{84.04\std{0.12}} \\
\midrule
\multicolumn{10}{c}{\textit{External Cohort}} \\
\midrule
LE-DUNet~\cite{lee2024synthesizing}           &18.05\std{0.08}  &19.64\std{3.74} &75.81\std{0.20}  &20.09\std{0.12}  &19.21\std{3.49}  &75.35\std{0.29}  
&23.31\std{0.27}  & 19.31\std{4.07} &72.86\std{0.50}  \\
LE-GAN~\cite{zhang2025language}     &14.28\std{0.06}  &20.67\std{3.75}  & 76.85\std{0.13} 
&14.09\std{0.05}  &20.88\std{2.92}  &76.54\std{0.11}  
& 15.54\std{0.07} &20.58\std{4.57}  &74.08\std{0.15}  \\
IL-CLDM~\cite{ou2024image}    &12.84\std{0.11}  &21.40\std{5.83}  &79.11\std{0.19} 
&12.98\std{0.06}  &21.22\std{4.65}  &78.74\std{0.17}    
&14.22\std{0.11} &21.42\std{4.43}  & 77.18\std{0.26} \\
LDM-GBM~\cite{leclercq2025pre}   &  10.55\std{0.06} &21.42\std{5.39}   &80.00\std{0.17} &10.86\std{0.05}&21.09\std{4.47}  &79.28\std{0.16}   
&12.63\std{0.11}  & 21.20\std{4.74} &77.23\std{0.27}  \\
MedWM~\cite{yang2025medical}   &10.48\std{0.08}  &21.90\std{6.12}  &80.18\std{0.18} 
&10.25\std{0.06}  &21.46\std{4.33}  &80.56\std{0.16}    
& 12.63\std{0.13} &21.67\std{4.07}  & 77.30\std{0.30} \\
TaDiff~\cite{liu2025treatment} &10.16\std{0.06}  & 22.05\std{5.58} &80.54\std{0.17} 
& 10.01\std{0.05}& 21.96\std{4.86} &81.13\std{0.15}     
&12.17\std{0.10}  &22.02\std{4.86}  & 77.75\std{0.26} \\

\cellcolor{cyan!10}\textbf{\methodname (Ours)} &\cellcolor{cyan!10}\,\,\,\textbf{9.55\std{0.07}} & \cellcolor{cyan!10} \textbf{22.58\std{5.43}} 
& \cellcolor{cyan!10} \textbf{81.46\std{0.18}}
& \cellcolor{cyan!10}\,\,\,\textbf{9.43\std{0.05}} & \cellcolor{cyan!10}\textbf{22.75\std{4.61}}& \cellcolor{cyan!10}\textbf{82.09\std{0.18}}
               & \cellcolor{cyan!10}\textbf{11.64\std{0.10}} & \cellcolor{cyan!10} \textbf{22.69\std{4.67}}& \cellcolor{cyan!10}\textbf{78.53\std{0.31}} \\
\shline
\end{NiceTabular*}
\end{table*}

\subsubsection{Evaluation Metrics}
Performance evaluation was conducted across two tasks. 
For {treatment planning}, we formulated the problem as multi-class classification~\cite{peng2025omnibrainbench} and compared predictions with clinical references using accuracy, F1-score, specificity, and precision. 
For {future MRI generation}, we assessed reconstruction fidelity~\cite{poli1993use,wang2004image} via voxel-wise normalized mean squared error (NMSE), peak signal-to-noise ratio (PSNR), and structural similarity index (SSIM).

\subsection{Comparison with SOTA methods}
We benchmarked our \methodname against SOTA methods on distinct tasks across both internal and external cohorts, assessing task-specific efficacy and cross-center generalizability.

\subsubsection{Treatment Planning Results}
To evaluate the treatment-planning capability of our \methodname, we compared it against four categories of strong baselines in \tabref{tab:compare_all_models}. The comparison included (\textsc{I}) open-source general-purpose vision-language models (Qwen3-VL 4B~\cite{bai2025qwen3vltechnicalreport}, InterVL3.5 4B~\cite{wang2025internvl3}, DeepSeek V3.2~\cite{liu2025deepseek}), (\textsc{II}) closed-source general models (Claude 4.5, Gemini 2.5 Pro~\cite{comanici2025gemini}, GPT 5.1~\cite{achiam2023gpt}), (\textsc{III}) medical domain-specific models (MedVLM-R1 2B~\cite{pan2025medvlm}, Hulu-Med 4B~\cite{jiang2025hulu}, Linshu 7B~\cite{xu2025lingshu}), and (\textsc{IV}) task-specific treatment-planning models (AliFuse~\cite{chen2024alifuse}, Hyperfusion~\cite{duenias2025hyperfusion}). All models were tasked with predicting the next-step treatment plan based on multimodal clinical and imaging contexts.
For the VLM baselines (types \textsc{I}–\textsc{III}), we bridged the architectural gap between 3D volumetric data and 2D vision encoders by processing the 3D MRI inputs as continuous sequences of 2D slices, conceptually similar to video inputs, whereas the task-specific models (type \textsc{IV}) processed the data natively using 3D encoders.

\begin{figure*}[!ht]
	\centering
\includegraphics[width=1.\linewidth]{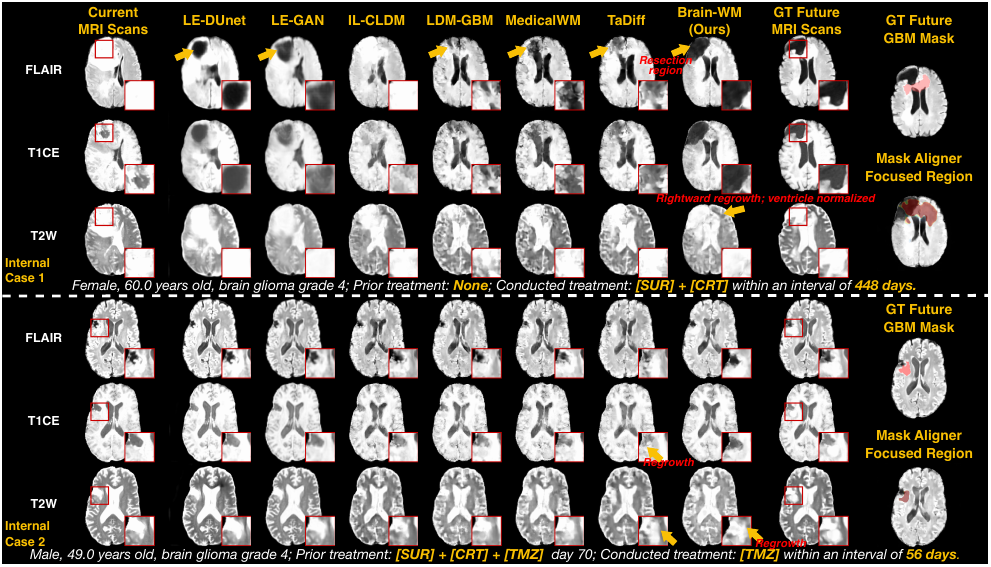}
 \caption{\revision{\textbf{Qualitative results of ours and other competing methods on the internal cohort.}}}
\label{fig:qua_internal}
\end{figure*}

\begin{figure*}[!ht]
	\centering
\includegraphics[width=1.\linewidth]{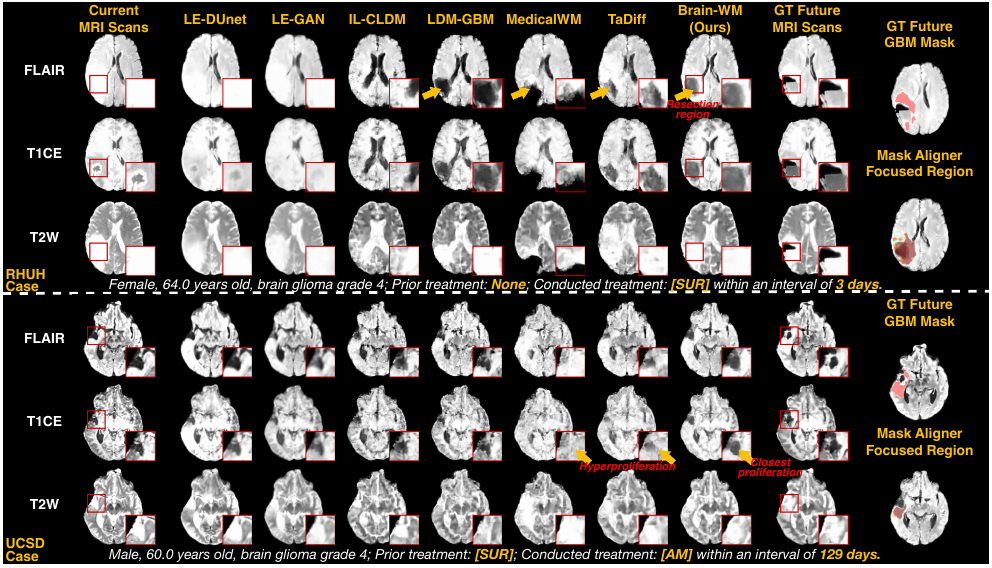}
 \caption{\revision{\textbf{Qualitative results of ours and other competing methods on the external cohort.}}}
\label{fig:qua_external}
\end{figure*}

Table~\ref{tab:compare_all_models} demonstrates that general-purpose VLMs, both open-source and closed-source, struggled with GBM treatment planning, confirming that broad instruction tuning is insufficient for specialized clinical reasoning.
While medical domain-specific models narrowed this gap by aligning with radiological and oncological semantics, they remained generic question-answering systems lacking the longitudinal reasoning required to track disease temporal progression.
We explicitly trained task-specific architectures, \ie~AliFuse and Hyperfusion, on our internal cohort. Despite this optimization, they remained constrained by their formulation as static classifiers.
Relying on shallow input-to-label mappings, they failed to capture spatiotemporal tumor dynamics, underscoring the limitations of snapshot-based discriminative methods.
Notably, while baselines like Qwen3-VL 4B and Claude 4.5 exhibited high precision, their low F1-scores betrayed a conservative bias that sacrificed sensitivity critical for oncological planning.
In contrast, \methodname\ jointly modeled treatment planning and future MRI generation within a shared latent space. 
Initialized from a multimodal foundation and refined via domain-specific training, it effectively captured complex GBM evolution, yielding superior robustness to distribution shifts across both internal and external cohorts.

\subsubsection{Future MRI Generation Results}
To assess \methodname{}'s ability to generate future MRIs, we comprehensively evaluated \methodname against SOTA 3D generation methods. Given the lack of standardized benchmarks for this specific 3D longitudinal task, we reproduced representative methods spanning DenseUNet (LE-DUNet~\cite{lee2024synthesizing}), GAN (LE-GAN~\cite{zhang2025language}), and diffusion families (IL-CLDM~\cite{ou2024image}, LDM-GBM~\cite{leclercq2025pre}, MedWM~\cite{yang2025medical}, TaDiff~\cite{liu2025treatment}). For fair comparison, we standardized clinical context integration using CLIP-based priors across all baselines. Among the diffusion-based approaches, conditioning strategies vary: LDM-GBM and MedWM explicitly utilized the current tumor mask as input to guide generation, whereas TaDiff employed it as implicit supervision by joint prediction of current and future masks.

As summarized in Table~\ref{tab:comp_gene} and Figs.~\ref{fig:qua_internal} and~\ref{fig:qua_external}, we evaluated the quantitative and qualitative performance of \methodname for future MRI generation. Focusing on the quantitative analysis, our study yielded five key observations: (1) Diffusion-based architectures established a clear performance advantage over UNet- and GAN-based baselines, driven by their superior capacity to model high-frequency details and complex pathological distributions. (2) Incorporating structural guidance proved pivotal, as structure-aware models, whether relying on explicit mask inputs (LDM-GBM, MedWM) or implicit auxiliary supervision (TaDiff), substantially outperformed the unconstrained baseline (IL-CLDM) across all evaluation metrics. (3) Crucially, implicit supervision achieved performance parity with explicit conditioning strategies, demonstrating that latent structural constraints can effectively guide synthesis without the prerequisite for inference-time tumor masks. (4) \methodname consistently achieved SOTA performance across both internal and external cohorts, utilizing its unified planning-imaging framework with  synergistic alignment constraints to secure the lowest NMSE and highest PSNR/SSIM, thereby maintaining robust generalization even under significant domain shifts where competing methods degrade. (5) Quantitative fidelity exhibited a distinct modality-dependent hierarchy, with T1CE yielding the highest quality followed by FLAIR, whereas T2W sequences proved the most challenging due to their heterogeneous tissue contrast.

Qualitative comparisons in Figs.~\ref{fig:qua_internal} and~\ref{fig:qua_external} demonstrate that while diffusion models improve texture over UNet and GANs, they lack the precise trajectory modeling achieved by the unified framework of our \methodname.
In internal case 1, although mask-aware methods (LDM-GBM, MedWM, TaDiff) recognized the resection intent, they exhibited substantial spatial deviations from the ground-truth cavity and failed to accurately predict the residual tumor regions.
In internal case 2, explicitly conditioned methods (LDM-GBM, MedWM) produced nearly static tumor volumes; conversely, the implicitly supervised TaDiff captured growth directionality on the FLAIR sequence but suffered from overestimated proliferation, hallucinating tissue invasion into the resection cavity, an artifact successfully eliminated by our \methodname.
Under domain shift in the external RHUH case, explicitly mask-conditioned baselines performed coarse, mechanical resections.
While TaDiff generated a plausible cavity, it failed to account for the clinical language context, predicting unreasonable tumor expansion despite the short 3-day interval; in contrast, \methodname produced a resection profile  highly consistent with the ground truth. 
Finally, in the external UCSD case, while MedWM and TaDiff modeled the tumor regrowth trend, MedWM exhibited hyper-aggressive proliferation that completely filled the resection cavity, and TaDiff showed a similar overestimation. Conversely, \methodname  achieved the closest alignment with the actual tumor evolution trajectory. 
Although exact voxel-level  reconstruction is intractable due to GBM heterogeneity and surgical variability, \methodname provides a reliable  anatomical reference  to assist clinicians in prospective assessment.

\subsection{Ablation Study}
This subsection first investigates the impact of individual components on two distinct tasks: treatment planning and future MRI generation, followed by an in-depth analysis of the synergistic multi-timepoint mask alignment (MM-Align). All ablation studies were conducted on the internal cohort.

\begin{table}[!tp]
\centering
\caption{
\textbf{Internal next-step treatment planning ablation results.}
}
\scriptsize
\label{tab:ablation_planning}
\begin{NiceTabular*}{1.\linewidth}{@{\extracolsep{\fill}}l|cccc}
\shline
{Model Variants}
&Accuracy $\uparrow$
&F1-Score $\uparrow$
&Specificity $\uparrow$
&Precision $\uparrow$ \\
\midrule
Planning-only
&85.9 &84.9 &91.8 &84.2 \\
\midrule
Unified model
&87.2 &88.6 &92.4 &88.4 \\
\;\:+Y-shaped MoT
&90.4 &89.1 &93.7 &89.3 \\
\cellcolor{cyan!10}\;\;\;+MM-Align (\textbf{Ours})
&\cellcolor{cyan!10}\textbf{91.5}
&\cellcolor{cyan!10}\textbf{90.4}
&\cellcolor{cyan!10}\textbf{95.8}
&\cellcolor{cyan!10}\textbf{90.7} \\
\shline
\end{NiceTabular*}
\end{table}

\begin{table*}[!tp]
\centering
\caption{
\textbf{Internal future MRI generation ablation results.}
}
\label{tab:ablation_generation}
\scriptsize
\begin{NiceTabular*}{\textwidth}{@{\extracolsep{\fill}}l|ccc|ccc|ccc}
\shline
\multirow{2}{*}{{Model Variants}}
&\multicolumn{3}{c|}{{FlAIR}}
&\multicolumn{3}{c|}{{T1CE}}
&\multicolumn{3}{c}{{T2W}} \\
  & NMSE $\downarrow$ & PSNR  $\uparrow$ & SSIM  $\uparrow$
  & NMSE $\downarrow$ & PSNR $\uparrow$ & SSIM  $\uparrow$
  & NMSE  $\downarrow$ & PSNR  $\uparrow$ & SSIM  $\uparrow$ \\
\midrule
Generation-only
&\,\,\,6.48\std{0.09} &24.24\std{3.26} &84.39\std{0.16}
&\,\,\,6.06\std{0.10} &24.39\std{3.19} &84.87\std{0.16}
&\,\,\,6.47\std{0.08} &24.35\std{3.21} &83.21\std{0.14} \\
\midrule
Unified model
&\,\,\,6.53\std{0.11} &23.91\std{3.11} &83.78\std{0.14}
&\,\,\,6.17\std{0.09} &24.18\std{3.58} &84.37\std{0.19}
&\,\,\,6.83\std{0.10} &23.96\std{2.75} &82.37\std{0.14} \\
\;\:+Y-shaped MoT
&\,\,\,6.24\std{0.09} &24.61\std{3.50} &84.74\std{0.16}
&\,\,\,5.84\std{0.11} &24.72\std{3.52} &85.24\std{0.15}
&\,\,\,6.26\std{0.07} &24.68\std{2.95} &83.55\std{0.13} \\
\cellcolor{cyan!10}\;\;\;+MM-Align (\textbf{Ours})
& \cellcolor{cyan!10} \,\,\,\textbf{5.93\std{0.11}}& \cellcolor{cyan!10}\textbf{25.05\std{3.41}} & \cellcolor{cyan!10} \textbf{85.24\std{0.14}}
& \cellcolor{cyan!10}
\,\,\,\textbf{5.46\std{0.09}} & \cellcolor{cyan!10}\textbf{25.29\std{3.54}} & \cellcolor{cyan!10}\textbf{85.81\std{0.18}}
& \cellcolor{cyan!10}\,\,\,\textbf{5.98\std{0.08}} & \cellcolor{cyan!10} \textbf{25.17\std{3.09}} & \cellcolor{cyan!10} \textbf{84.04\std{0.12}} \\
\shline
\end{NiceTabular*}
\end{table*}

\subsubsection{Ablation on Proposed Contributions}
We conducted a systematic ablation study to demonstrate the effectiveness of each proposed component in modernizing the task-specific baseline architecture into \methodname. As detailed in~\tabref{tab:ablation_planning} for treatment planning and~\tabref{tab:ablation_generation} for future MRI generation, we started with single-task baselines and successively integrated the basic unified formulation, the Y-shaped MoT, and finally the synergistic multi-timepoint mask alignment strategy.

For the treatment planning task, transitioning from the planning-only baseline to the unified model, \ie~effectively introducing the generation task for the same architecture, yielded immediate performance gains.
This finding indicated that the additional future MRI generation facilitated the learning of GBM spatiotemporal dynamics by enforcing tumor-aware feature learning and promoting inter-task synergy.
The Y-shaped MoT further boosted performance by structurally decoupling task-shared from task-specific learning.
Finally, introducing multi-timepoint alignment provided additional gains, attributed to tumor-aware supervision that explicitly constrained feature semantics to better capture longitudinal tumor evolution.

In contrast to the immediate gains observed in planning, the ablation results for future MRI generation exhibited an asymmetric pattern. Starting from a generation-only baseline, the direct addition of the treatment planning task slightly deteriorated reconstruction quality across all modalities, evidenced by increased NMSE and decreased PSNR/SSIM.
This indicated task interference, likely due to the competing needs for semantic abstraction (planning) versus voxel-level reconstruction (generation), when parameters were fully shared between conflicting objectives.
However, by disentangling task-specific feature refinement through dedicated FFNs while retaining a shared attention backbone, the Y-shaped MoT mitigated task interference; consequently, imaging performance not only recovered but also surpassed the generation-only baseline on FLAIR, T1CE, and T2W.
Finally, aligning shared features via multi-timepoint masks yielded consistent improvements.

Overall, ablations showed a distinct dynamic: while planning benefits inherently from the generation task, generation requires Y-shaped MoT structural decoupling and alignment constraints to convert  interference into constructive synergy.

\begin{figure}[!t]
	\centering
\includegraphics[width=1.\linewidth]{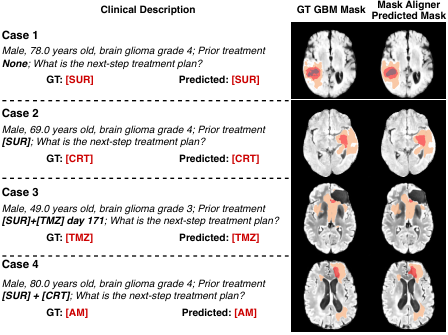}
\caption{\textbf{Qualitative visualization of the masks predicted by the lightweight mask aligner during treatment planning.}}
\label{fig:treat_visual}
\end{figure}

\subsubsection{Further Analysis of MM-Align} 
To qualitatively validate the efficacy of the MM-Align, we presented mask visualization results for treatment planning in \figref{fig:treat_visual} and future MRI generation in Figs.~\ref{fig:qua_internal}--\ref{fig:qua_external}.
For treatment planning, \methodname produced morphologically consistent segmentation masks, demonstrating its capacity to comprehend tumor topology before deriving appropriate treatment plans (Fig.~\ref{fig:treat_visual}).
Note that MM-Align solely employed a lightweight mask aligner operating on intermediate features that, while not designed for voxel-accurate segmentation, produced anatomically meaningful tumor-aware priors that synergistically supported both treatment planning and future MRI generation.
Regarding the future MRI generation task, we visualized the model's attentional focus by aggregating predicted segmentation labels weighted by inference noise (Figs.~\ref{fig:qua_internal}--\ref{fig:qua_external}).
The results indicated that the predicted masks generally aligned well with the manual annotations and were spatially consistent with the generated tumor regions, confirming that MM-Align effectively guided the generation process.
Notably, in the two surgical resection cases (Internal case 1 and RHUH case), the alignment exhibited larger deviations from the ground truth, reflecting the challenges of future prediction and subjective variability in surgical operations.

\begin{figure}[!t]
\centering
\includegraphics[width=1\columnwidth]{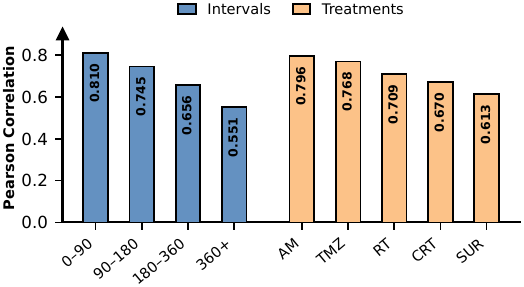} 
\caption{\textbf{Correlation between  the MM-Align focused-region volume in future
MRI generation and the ground-truth tumor volume.}}
\label{fig:correlation_analysis}
\end{figure}

\revision{Fig.~\ref{fig:correlation_analysis} presents the Pearson correlation ($r$) results between the MM-Align focused-region volume during future MRI generation and the ground-truth tumor volume, stratified by prediction interval and treatment type. Over time,} the \revision{correlation $r$ exhibited a monotonic decay from $0.810$ (0--90 days) to $0.551$ ($>360$ days), reflecting the accumulation of aleatoric uncertainty in stochastic tumor dynamics.} 

\revision{Across treatment types, the model captured naturalistic progression under active monitoring (AM, $r=0.796$) and the systemic response pattern associated with temozolomide (TMZ, $r=0.768$). In contrast, performance declined as interventions became more complex. Specifically, the lower correlation for concurrent chemoradiotherapy (CRT, $r=0.670$) reflected the difficulty of disentangling mixed radiographic signals and confounders such as pseudoprogression. Finally, the surgery group exhibited the weakest correlation ($r=0.613$), primarily due to abrupt postoperative anatomical changes and inter-operator variability in resection extent, which weaken the linkage between preoperative imaging-derived volumes and subsequent tumor burden.}

\section{Conclusion}
\methodname advances GBM trajectory simulation by adopting a unified world-modeling framework that couples tumor evolution with clinical decision-making.
By seamlessly integrating autoregressive next-step treatment planning and flow-based future MRI generation, our framework empowers clinicians with a risk-free sandbox to explore counterfactual scenarios and optimize patient healthcare. We acknowledge that the current version presents a simplified modeling of intervention tasks, prioritizing macroscopic tumor evolution over fine-grained radiotherapeutic dosimetry. Future work will seek to address this limitation by incorporating radiation dose maps and molecular biomarkers to further enrich the patient-specific digital twin.
Moving forward, we plan to prospectively evaluate \methodname  to assess its clinical utility as a human-in-the-loop decision-support tool for precision oncology.

\newpage
\bibliographystyle{IEEEtran.bst}
\bibliography{ref}

\end{document}